\documentclass[11pt]{article}
\usepackage{eacl2017}
\usepackage{times}
\usepackage{url}
\usepackage{latexsym}
\usepackage{color}
\usepackage{amsfonts}
\usepackage[boxruled]{algorithm2e}
\usepackage{graphicx}
\usepackage{paralist}

\eaclfinalcopy 

\title{Vicinity-Driven Paragraph and Sentence Alignment \\ for Comparable Corpora}

\author{
Gustavo Henrique Paetzold \and Lucia Specia \\
  Department of Computer Science \\ University of Sheffield, UK \\
  {\tt \{g.h.paetzold,l.specia\}@sheffield.ac.uk} \\ 
	}

\date{}

\begin{document}
\maketitle
\begin{abstract}
Parallel corpora have driven great progress in the field of Text Simplification. However, most sentence alignment algorithms either offer a limited range of alignment types supported, or simply ignore valuable clues present in comparable documents. We address this problem by introducing a new set of flexible vicinity-driven paragraph and sentence alignment algorithms that 1-N, N-1, N-N and long distance null alignments without the need for hard-to-replicate supervised models.
\end{abstract}

\section{Introduction}
\label{sec:intro}
Early Text Simplification (TS) approaches relied heavily on manually-crafted simplification rules \cite{chandrasekar1996motivations,Siddharthan} due to a lack of resources from which to automatically learn or extract them. Complex-to-simple parallel corpora effectively changed the nature of contributions pertaining to TS upon their arrival.

The most widely used resource of this kind is the Wikipedia and Simple Wikipedia parallel corpus \cite{Kauchak2006}. Using this corpus, authors have devised Syntactic Simplification approaches that learn simplification rules from aligned sentences \cite{Siddharthan2006,Woodsend2011}, Machine Translation approaches that translate from complex to simple text \cite{Zhu2010,Bach2011}, Lexical Simplifiers that extract complex-to-simple word correspondences from word alignments \cite{Biran2011,Horn2014}, and even hybrid lexico-syntactic simplifiers \cite{Paetzold2013,Feblowitz2013}. Other examples of parallel corpora that have been used in TS are the medical-domain corpus of \cite{Deleger2009}, and the newly introduced Newsela corpus \cite{xu2015problems}.

Most of these corpora come, however, as a set of complex-to-simple articles aligned at document level. In order for them to be used by Machine Translation systems, Lexical Simplification strategies and Tree Transduction frameworks, for example, these documents need to be aligned at sentence level.

Various strategies have been created to address this problem. \newcite{barzilay2003sentence} present an approach that first aligns paragraphs with sophisticated clusterers and classifiers learned from manually annotated instances, then produces sentence alignments through a dynamic programming algorithm that finds an alignment path between the sentences in aligned paragraphs. \newcite{Coster11} use the same sentence alignment approach, but opt instead for a much simpler paragraph alignment algorithm: they simply align any pair of paragraphs of which the TF-IDF similarity is higher than a manually set threshold of 0.5. Similarly, \newcite{Bott2011} also make use of an elaborate, difficult to replicate supervised approach in order to learn the likelihood of two portions of text being aligned. \newcite{smith2010extracting} bypass the step of paragraph alignment and use instead a supervised Conditional Random Fields ranker for sentence alignment directly. 

Although these algorithms have been proven useful, they do not exploit the comparability between the aligned documents' paragraphs and sentences, hence the need for elaborate supervised classifiers and rankers. The sentence aligner used by both \newcite{barzilay2003sentence} and \newcite{Coster11} suffers from yet another limitation: it does not allow for multiple consecutive sentence skips, which can lead to incorrect alignments in scenarios where the number of sentences in a pair of aligned paragraphs is not similar.

In this paper, we introduce a new set of paragraph and sentence alignment algorithms designed to address these limitations.

\section{Clues in Comparable Corpora}
\label{sec:newselacorpus}

By inspecting a handful of documents from the Wikipedia-Simple Wikipedia and the Newsela corpus, we noticed that they are divided in numerous paragraphs, and that the order with which the information is presented is consistent throughout different versions of the same article. This means that, most of the time, one can safely assume that, if the $i$th and $j$th paragraph of a pair of documents is aligned, then the next alignment $\left ( x, y \right )$ will be such that $x\! \geq\! i$ and $y\! \geq\! j$.

Since paragraphs offer valuable alignment clues, we believe that algorithms that bypass paragraph alignment, such as that of \newcite{smith2010extracting}, are not suitable for this task. We also noticed that the order of the information within aligned paragraphs is also consistent, but that often times aligned paragraphs will have a noticeably large disparity in number of sentences they have. Consequently, sentence alignment algorithms that do not allow for multiple consecutive skips, such as that of \newcite{barzilay2003sentence}, may not be suitable for this task either.

\newcite{xu2015problems} produced sentence alignments for the Newsela corpus by simply calculating the Jaccard similarity score of each pair of sentences in a set of documents, then selecting any pairs that achieve a similarity score above a manually set threshold. This approach, however, suffers from several limitations also, since it does not exploit the aforementioned observations in any capacity, nor does it allow for N-1 or N-N alignments.

We address this problem by introducing new, flexible paragraph and sentence alignment algorithms that better document-aligned corpora. We describe them in what follows.

\section{Paragraph Alignment Algorithm}

In order to produce paragraph alignments, we employ a search algorithm that allows 1-1, 1-N, N-1, N-N and null alignments (i.e. unaligned paragraphs between alignments).

Algorithm~\ref{algparag} exploits a matrix $M\! \in\! \mathbb{R}^{\left \| D_{1} \right \|\times \left \| D_{2} \right \|}$ where $\left \| D_{1} \right \|$ represents the number of paragraphs in a document $D_{1}$, $\left \| D_{2} \right \|$ the number of paragraphs in a document $D_{2}$, and $M\! \left (i, j \right )$ the similarity between the $i$th paragraph in $D_{1}$ and the $j$th paragraph in $D_{2}$. As a similarity metric, we use the maximum TF-IDF cosine similarity between all possible pairs of sentences in $\left [ i, j \right ]$ i.e. the similarity between two paragraphs is equal to that of the most similar pair of sentences in them. We choose this metric due to the fact that, even though the sentences in equivalent paragraphs of documents with different reading levels are usually very distinct in both form and vocabulary, often times paragraphs will share at least one very similar (or even identical) sentence between them, which is a strong indicator of a good paragraph alignment. These similar sentences are often so because of long quotes from subjects interviewed for the news articles.

The goal of Algorithm~\ref{algparag} is to find a path $A\! \in\! \mathbb{Z}^{\left \| D_{1} \right \|\times \left \| D_{2} \right \|}$ where $A \! \left ( i, j  \right )\! =\! 1$ if there is an alignment between paragraphs $\left [ i, j  \right ]\in D_{1}\! \times\! D_{2}$, and $A \! \left ( i, j  \right ) \! =\! 0$ otherwise. The search for path $A$ starts from the assumption that $A \! \left ( 1, 1  \right ) = 1$, or in other words, the first paragraph of $D_{1}$ and $D_{2}$ are aligned.
We exploit this assumption because the first paragraph in most document-aligned corpora available refer to the articles' titles. The algorithm then initializes a control set of coordinates $\left [ c_{x}, c_{y} \right ] = \left [ 1, 1\right ]$ that represents the point from which to search for the next alignment in $A$. The next alignment is then searched for in the first vicinity V1, which represents 1-1, 1-N and N-1 alignments. If there is not a paragraph pair $\left [ n_{x}, n_{y} \right ]$ in V1 such that $M\! \left (n_{x}, n_{y} \right )\! \geq \! \alpha$ then the next alignment is searched for in the second vicinity V2, which represents single paragraph skips. If no pair $\left [ n_{x}, n_{y} \right ]$ in V2 has $M\! \left ( n_{x}, n_{y} \right )\! \geq \! \alpha$, then the algorithm searches for the the next pair $\left [ n_{x}, n_{y} \right ]$ with $M\! \left (n_{x}, n_{y} \right )\! \geq \! \alpha$ with the shortest euclidean distance to $\left [ c_{x}, c_{y} \right ]$ in the third and final vicinity V3, which represents long-distance paragraph skips. Finally, the update $\left [ c_{x}, c_{y} \right ]=\left [ n_{x}, n_{y} \right ]$ is made, and the alignment $A \! \left ( c_{x}, c_{y}  \right ) \! =\! 1$ is added. This process is repeated until there are no more paragraphs left to be aligned, or until the algorithm finds no suitable $\left [ n_{x}, n_{y} \right ]$ to follow current alignment $\left [ c_{x}, c_{y} \right ]$ in vicinity V3. Notice that, although we use only three vicinities, the algorithm can be easily adapted to support as many distinct vicinities as suitable. After $A$ is found, the paragraphs in all 1-N, N-1 and N-N alignments are concatenated so that our sentence alignment algorithm can more easily search for equivalent sentences in them.

\begin{algorithm}[htpb]
\small
\SetInd{0.5em}{0.5em}
\caption{Paragraph Alignment\label{algparag}}

	input: $M\!$, $D_{1}$, $D_{2}$, $\alpha$\;
	output: $A$\;
	
	\BlankLine
	
	$A \leftarrow zeros\left ( \left \| D_{1} \right \|, \left \| D_{2} \right \| \right )$\;
	$\left [ c_{x}, c_{y} \right ]\leftarrow \! \left [ 1,1 \right ]$\;
	
	\BlankLine
	
	\While{$\left [ c_{x}, c_{y} \right ]\neq nil$}{
		$A\! \left ( c_{x}, c_{y} \right ) \leftarrow 1$\;
		
		\BlankLine
		
		$\textup{V1} \leftarrow \left \{ \left [ c_{x}, c_{y}\!+\!1 \right ],\left [ c_{x}\!+\!1, c_{y} \right ],\left [ c_{x}\!+\!1, c_{y}\!+\!1 \right ] \right \}$\;
		$\textup{V2} \leftarrow \left \{ \left [ c_{x}\!+\!2, c_{y}\!+\!1 \right ],\left [ c_{x}\!+\!1, c_{y}\!+\!2 \right ] \right \}$\;
		$\textup{V3} \leftarrow \left \{\left [x,y \right ] \mid l_{x}\geq x\geq c_{x}\, \,  \textup{and} \, \, l_{y}\geq y\geq c_{y}\right \}$\;
		
	\BlankLine
	
	$\left [ n_{x}, n_{y} \right ]\leftarrow \left [ x,y \right ]$ with best $M\! \left ( x,y \right )$ in $\textup{V1}$\;
		\If{$M\! \left ( n_{x}, n_{y} \right )< \alpha $}{
			$\left [ n_{x}, n_{y} \right ]\leftarrow \left [ x,y \right ]$ with best $M\! \left ( x,y \right )$ in $\textup{V2}$\;
			\If{$M\! \left ( n_{x}, n_{y} \right )< \alpha $}{
				$\left [ n_{x}, n_{y} \right ]\leftarrow \left [ x,y \right ]$ with $M\! \left ( x,y \right )\! \geq\! \alpha$ in $\textup{V3}$ \\ with shortest euclidean distance to $\left [ c_{x}, c_{y} \right ]$\;
			}
		}
		
	\BlankLine
		
		$\left [ c_{x}, c_{y} \right ]\leftarrow \left [ n_{x}, n_{y} \right ]$\;
	}
\end{algorithm}

\section{Sentence Alignment Algorithm}

Our sentence alignment algorithm employs the same principles behind our paragraph alignment algorithm: it also allows for 1-1, 1-N, N-1 and null alignments, and works under the assumption that N-N alignments can be inferred from consecutive 1-1 alignments.

Algorithm~\ref{algsent} exploits a matrix $M\! \in\! \mathbb{R}^{\left \| P_{1} \right \|\times \left \| P_{2} \right \|}$ where $\left \| P_{1} \right \|$ represents the number of sentences in a paragraph $P_{1}$, $\left \| P_{2} \right \|$ the number of sentences in a paragraph $P_{2}$, and $M\! \left (i, j \right )$ the similarity between the $i$th sentence in $P_{1}$ and the $j$th sentence in $P_{2}$. As a metric, we use TF-IDF cosine similarity.

In search of path $A\! \in\! \mathbb{Z}^{\left \| P_{1} \right \|\times \left \| P_{2} \right \|}$, our algorithm first finds the initial alignment point $\left [ c_{x}, c_{y} \right ]$ with $M\! \left (c_{x}, c_{y} \right )\! \geq \! \alpha$ with the shortest euclidean distance to $\left [ 0, 0 \right ]$. Notice that this algorithm cannot exploit the assumption that the first sentences in a pair of paragraphs will be aligned. The next alignment $\left [ n_{x}, n_{y} \right ]$ is then searched for in the immediate vicinity V1. If $M\! \left ( n_{x}, n_{y} \right )$ is smaller than a minimum $\alpha$, then it searches for the the next pair $\left [ n_{x}, n_{y} \right ]$ with $M\! \left ( n_{x}, n_{y} \right )\! \geq \! \alpha$ with the shortest euclidean distance to $\left [ c_{x}, c_{y} \right ]$ in vicinity V2. Otherwise, if $\left [ n_{x}, n_{y} \right ] = \left [ c_{x}\! +\! 1, c_{y}\! +\! 1 \right ]$, then the new 1-1 alignment is added to $A$. However, if a 1-N ($\left [ n_{x}, n_{y} \right ] = \left [ c_{x}, c_{y}\! +\! 1 \right ]$) or N-1 ($\left [ n_{x}, n_{y} \right ] = \left [ c_{x}\! +\! 1, c_{y} \right ]$) alignment is found, then the algorithm performs a secondary loop in order to find the size of $N$. $N$ is incremented until the accumulated similarity for the concatenated sentences given the current size of $N$ is smaller than the similarity for $N-1$ minus a slack value $\beta$, or until the similarity of the adjacent diagonal ($M\! \left ( n_{x}\! +\! 1, n_{y}\! :\! n_{y}\! +\! N \right )$ or $M\! \left ( n_{x}\!:\! n_{x}\! +\! N, n_{y}\! +\! 1 \right )$) is larger than the accumulated similarity. This process is repeated until there are no more paragraphs left to be aligned, or until the algorithm finds no suitable new alignment candidates.

\begin{algorithm}[t]
\small
\SetInd{0.5em}{0.5em}
\caption{Sentence Alignment\label{algsent}}

	input: $M\!$, $P_{1}$, $P_{2}$, $\alpha$, $\beta$\;
	output: $A$\;
	
	\BlankLine
	
	$A \leftarrow zeros\left ( \left \| P_{1} \right \|, \left \| P_{2} \right \| \right )$\;
	$\left [ c_{x}, c_{y} \right ]\leftarrow \left [ x,y \right ]$ with $M\! \left ( x,y \right )\! \geq\! \alpha$ \\ with shortest euclidean distance to $\left [ 0, 0 \right ]$\;
	
	\BlankLine
	
	\If{$\left [ c_{x}, c_{y} \right ] \neq nil$}{
		$A\! \left ( c_{x}, c_{y} \right )\leftarrow 1$\;
	}
	
	\BlankLine
	
	\While{$\left [ c_{x}, c_{y} \right ]\neq nil$}{
	
		$\textup{V1} \leftarrow \left \{ \left [ c_{x}, c_{y}\!+\!1 \right ],\left [ c_{x}\!+\!1, c_{y} \right ],\left [ c_{x}\!+\!1, c_{y}\!+\!1 \right ] \right \}$\;
		$\textup{V2} \leftarrow \left \{\left [x,y \right ] \mid l_{x}\geq x\geq c_{x}\, \,  \textup{and} \, \, l_{y}\geq y\geq c_{y}\right \}$\;
		
	\BlankLine
	
	$\left [ n_{x}, n_{y} \right ]\leftarrow \left [ x,y \right ]$ with best $M\! \left ( x,y \right )$ in $\textup{V1}$\;
	
	\uCase{$M\! \left ( n_{x}, n_{y} \right )< \alpha $}{
		$\left [ n_{x}, n_{y} \right ]\leftarrow \left [ x,y \right ]$ with $M\! \left ( x,y \right )\! \geq\! \alpha$ in $\textup{V2			}$ \\ with shortest euclidean distance to $\left [ c_{x}, c_{y} \right ]$\;
		$\left [ c_{x}, c_{y} \right ]\leftarrow \left [ n_{x}, n_{y} \right ]$\;
		$A\! \left ( c_{x}, c_{y} \right )\leftarrow 1$\;
	}
	
	\uCase{$\left [ n_{x}, n_{y} \right ] = \left [ c_{x}\!+\!1, c_{y}\!+\!1 \right ]$}{
		$\left [ c_{x}, c_{y} \right ]\leftarrow \left [ n_{x}, n_{y} \right ]$\;
		$A\! \left ( c_{x}, c_{y} \right )\leftarrow 1$\;
	}
	
	\uCase{$\left [ n_{x}, n_{y} \right ] = \left [ c_{x}, c_{y}\!+\!1 \right ]$}{
		$A\! \left ( n_{x}, n_{y} \right )\leftarrow 1$\;
		$N\leftarrow 1$\;
		\While{$M\! \left ( n_{x}, n_{y}\! :\! n_{y}\!\!  +\! \! N \right )\!\! >\!\! M\! \left ( n_{x}, n_{y}\! :\! n_{y}\! +\! N\!\! -\!\! 1 \right )\!\! -\!\!\beta$ \textup{and} $M\! \left ( n_{x}, n_{y}\! :\! n_{y}\!\! +\!\! N \right )\!\! >\!\! M\! \left ( n_{x}\!\! +\!\! 1, n_{y}\! :\! n_{y}\!\! +\!\! N \right )$}{
			$A\! \left ( n_{x}, n_{y}\! +\! N \right )\leftarrow 1$\;
			$N\leftarrow N\! +\! 1$\;
		}
		$\left [ c_{x}, c_{y} \right ]\leftarrow \left [ n_{x}, n_{y}\! +\! N \! -\! 1 \right ]$\;
	}
	
	\Case{$\left [ n_{x}, n_{y} \right ] = \left [ c_{x}\!+\!1, c_{y} \right ]$}{
		$A\! \left ( n_{x}, n_{y} \right )\leftarrow 1$\;
		$N\leftarrow 1$\;
		\While{$M\! \left ( n_{x}\! :\! n_{x}\!\! +\!\! N, n_{y} \right )\!\! >\!\! M\! \left ( n_{x}\! :\! n_{x}\!\! +\!\! N\!\! -\!\! 1, n_{y} \right )\!\! -\!\!\beta$ \textup{and} $M\! \left ( n_{x}\! :\! n_{x}\!\! +\!\! N, n_{y} \right )\!\! >\!\! M\! \left ( n_{x}\!:\! n_{x}\!\! +\!\! N, n_{y}\!\! +\!\! 1 \right )$}{
			$A\! \left ( n_{x}\! +\! N, n_{y} \right )\leftarrow 1$\;
			$N\leftarrow N\! +\! 1$\;
		}
		$\left [ c_{x}, c_{y} \right ]\leftarrow \left [ n_{x}\! +\! N \! -\! 1, n_{y} \right ]$\;
	}
	}
\end{algorithm}

\section{Final Remarks}
We presented a new set of vicinity-driven paragraph and sentence alignment algorithms for document-aligned corpora. Our algorithms addresses the limitations of previous approaches by exploiting clues in comparable documents that are often neglected. Unlike many early strategies, our algorithms allow for 1-N, N-1, N-N and long-distance null alignments.

In the future, we aim to conduct both intrinsic and extrinsic performance comparisons between our algorithms and earlier approaches. We also aim to employ our algorithms in the creation of new paragraph and sentence-aligned corpora to be made available to the public.


\bibliography{references}
\bibliographystyle{eacl2017}

\end{document}